\documentclass[journal,twoside]{IEEEtran}
\usepackage{amsmath}
\usepackage{amssymb}
 \usepackage{amsthm}
 \usepackage{hyperref}
 \usepackage{subfigure}
 \usepackage{tikz}
 \usetikzlibrary{arrows,positioning,shadows}
 \usepackage{standalone}
 \usepackage{color,soul}
 \usepackage{tabularx,ragged2e,booktabs,caption}
 \usepackage{flexisym}
 \usepackage{multirow}
\usepackage[ruled,vlined]{algorithm2e}

\title{DFraud$^3$ - Multi-Component Fraud Detection free of Cold-start}

\begin{document}

\title{DFraud$^3$ - Multi-Component Fraud Detection free of Cold-start}
\author{Saeedreza Shehnepoor*, Roberto Togneri, Wei Liu, Mohammed Bennamoun
\thanks{S. Shehnepoor (*corresponding author) is with the University of Western Australia, Perth, Australia.
R. Togneri is with the University of Western Australia, Perth, Australia.
M. Buneman is with the University of Western Australia, Perth, Australia.
W. Liu is with the University of Western Australia, Perth, Australia.
emails: \{saeedreza.shehnepoor@research.uwa.edu.au, roberto.togneri@uwa.edu.au, wei.liu@uwa.edu.au, mohammed.bennamoun@uwa.edu.au.\}}

} 
\maketitle
\begin{abstract}
Fraud review detection is a hot research topic in recent years. The Cold-start is a particularly new but significant problem referring to the failure of a detection system to recognize the authenticity of a new user. 
State-of-the-art solutions employ a translational knowledge graph embedding approach 
(\textit{TransE}) to model the interaction of the components of a review system. 
However, these approaches suffer from the limitation of \textit{TransE} in handling N-1 relations and the narrow scope of a single classification task, i.e., detecting fraudsters only. 
In this paper, we model a review system as a Heterogeneous Information Network (HIN) which enables a unique representation to every component and performs graph inductive learning on the review data through aggregating features of nearby nodes. 
HIN with graph induction helps to address the camouflage issue (fraudsters with genuine reviews) which has shown to be more severe when it is coupled with cold-start, i.e., new fraudsters with genuine first reviews. 
In this research, instead of focusing only on one component, detecting either fraud reviews or fraud users (fraudsters), vector representations are learnt for each component, enabling multi-component classification. In other words, we are able to detect fraud reviews, fraudsters, and fraud-targeted items, thus the name of our approach DFraud$^3$.
DFraud$^3$ demonstrates 
a significant accuracy increase of 13\% over the state of the art on Yelp. 


\end{abstract}

\section{Introduction}
\label{sec:intro}

Reading through online reviews before making a purchase is increasingly a common practice of consumers. Studies \cite{YelpRate2018} show that a rating increase of 1-star in Yelp  may lead to a 5-9\% in a surge of increase for a restaurant. The financial implications of online reviews are becoming significant which incentivise some businesses to pay imposters to write fake comments, i.e. fraud reviews, to either promote one's own business or defame competitors. Experts estimate that between 9\% to 40\% of reviews in Amazon are fraud \cite{YelpRate2018}. 


Given the challenging nature of fraud review detection, even humans can only achieve an accuracy close to a random guess. 
It is, therefore, not surprising to see a surge of research effort in this area. 
To ensure a clear discussion of the research done in this area, let us model a review platform as a triple $\langle$review, user, item$\rangle$ where a review is written by a user for an item.
Fraud detection algorithms typically rely on historical data to extract behavioral patterns of users, which have shown to be more effective than linguistic features \cite{Shebuit2015,Shehnepoor2017} for fraud review detection. 
A key problem resulting from the reliance on historical data in such fraud detection systems is the phenomenon of \textit{cold-start}. Cold-start refers to the failure of a detection system to recognize the authenticity of a new user $u$ given the first review $r$ on an item $i$, since there is no historical information about that user. Furthermore, detecting fraud reviews and fraudsters may take time, and even when they are detected, the fraud reviews have already had their negative impacts.
The situation is exacerbated when new fraudsters apply \textit{the camouflage strategy} in their first reviews. 

Camouflage \cite{Shebuit2015,Shehnepoor2017,Hooi:2016:FBG:2939672.2939747,Hooi:2017:GFD:3119906.3056563} refers to the act of writing genuine reviews by fraudsters to hide their true identity and mask their traces. As a result of this act a fraudster gains the trust of other people before writing his/her first fraud review. Surprisingly, most fraudsters start their activity with genuine reviews, in order to cover up their true identity. In fact, statistics on the widely used Yelp dataset show that 62.18\% of fraudsters (1319 users out of 2121 camouflaged users) started their activity by writing genuine reviews. Intuitively, information from other components can be used to predict the probability of camouflage behaviors. For example, a review from a new user for an item frequently targeted by fraudsters is more likely to be a fraud \cite{Shebuit2015}. Hence, multi-component classification to classify reviews into genuine or not, users into fraudsters or not, and items into targeted or not, plays a very important role in handling cold-start, even when camouflage is employed by new fraudsters. 

\begin{figure}[ht]
  \includegraphics[width=\linewidth]{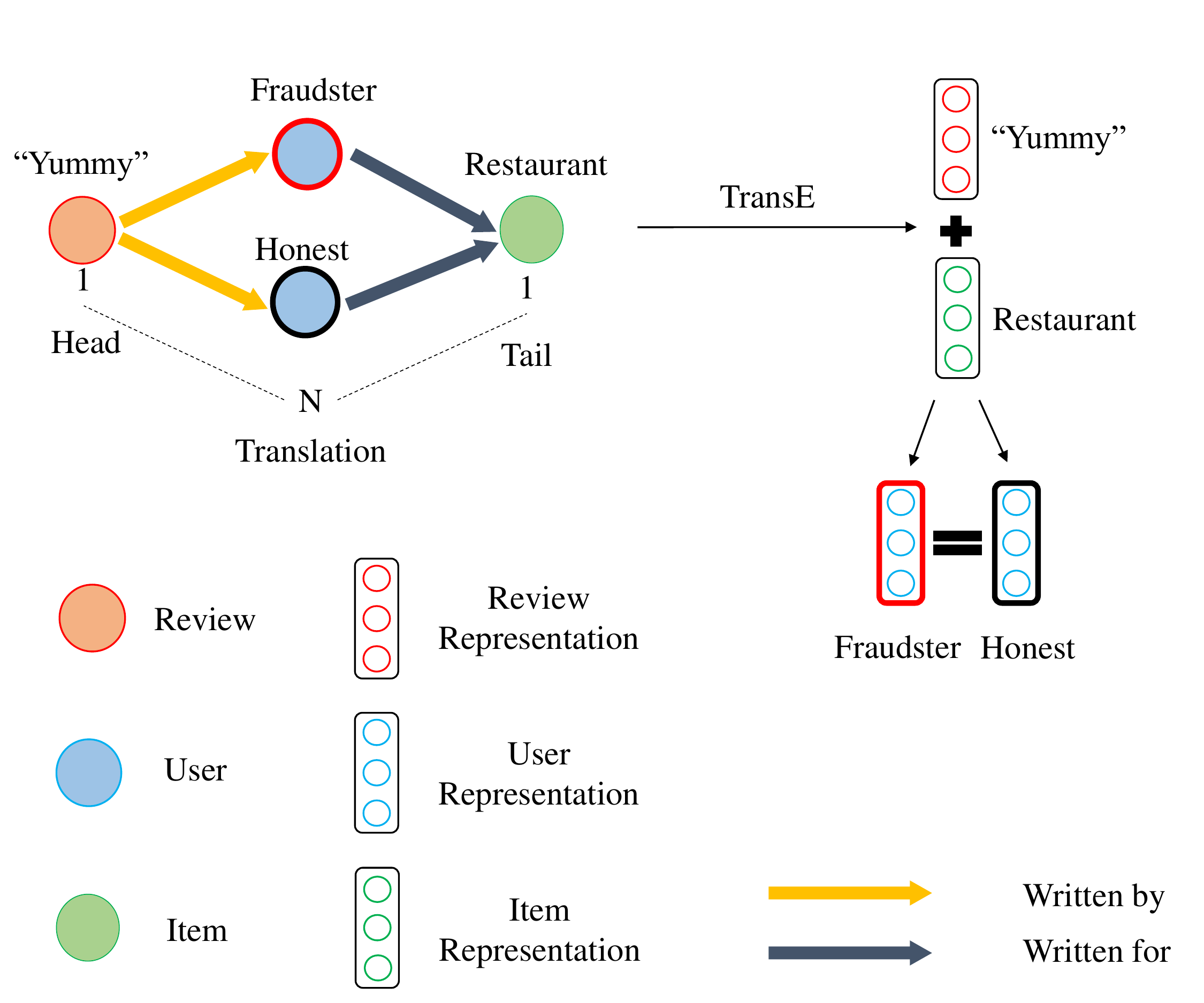}
  \caption{\textit{TransE} limitation in handling 1-N-1 relations, resulting in the same vector representations for both honest and fraudster user, which renders the user embeddings ineffective in differentiating fraudsters from genuine users.}
    \label{fig:1-N-1-diagram}
\end{figure}

Recent attempts at the cold-start problem \cite{wang2017handling,you2018attribute} adopted a knowledge graph embedding approach to model the relation between three components, namely, review, user, and item.
To learn their respective vector representations, Wang \textit{et al.} \cite{wang2017handling} and You \textit{et al.} \cite{you2018attribute} adopted the \textit{TransE}~\cite{Bordes:2013:TEM:2999792.2999923} embedding model, attempting to jointly learn the salient features representing each of the three components. 
However, despite \textit{TransE}'s simplicity and effectiveness in capturing multiple relations, its well known limitation is that it only works for 1-to-1 but not 1-to-N nor N-to-1 relations~\cite{Huynh:2018:TBF:3184558.3191616}. This is a significant drawback for the fraud review detection domain, because it is quite common for the same user to write similar reviews to different services.
Take the Yelp dataset 
for instance, 5.56\% of users (5,034 out of 90,177) wrote similar reviews (reviews like ``Yummy")
for different items. In \textit{TransE} parlance, for these users, one review (same content) is translated through one or more users to describe multiple items, thus exhibiting a N-1 relationship. In addition, 6.62\% of items (334 out of 5,044) have similar reviews (e.g., ``Great Steaks" or ``Awesome") from different users, reflecting the 1-N-1 relation (same review, different users, same item) as illustrated in Fig \ref{fig:1-N-1-diagram}. 
This limitation causes multiple users modelled as relations in \textit{TransE} to have identical vector representations, 
as was also observed by~\cite{DBLP:journals/corr/abs-1801-08641}.

Accordingly, modeling users as a separate component is fundamental in obtaining a useful representation for each user, which \textit{TransE} fails to achieve. Moreover, the camouflage problem is neglected, despite that it may significantly affect the performance of fraud review detection systems when coupled with cold-start. 
This calls for a better representation learning model with the ability to represent the intrinsic multi-relations between components, and to help spot the fraudsters when they start writing either fraud reviews from the beginning or genuine reviews to camouflage themselves.

Heterogeneous Information Networks (HIN) have been demonstrated to be suitable 
when it comes to gathering information from interconnected components \cite{Shehnepoor2017}.  
In this research, to address the limitations of \textit{TransE}, we choose to use an HIN as a more natural model for social review platform representation.
Contrary to the random vector strategy for network component initialization, in this research, we argue the importance of an appropriate component vector representation. Based on the theory of Collective Intelligence \cite{Malone2010,malone-harnessing}, aggregations of reviews are used to characterise each component in the network. In other words, a review's vector representation is the Sum of Word Embedding (SoWE) of all tokens in the review; a user's vector representation is then the SoWE of all reviews written by this user; and an item is the SoWE of all reviews about this item. The SoWE are further fine-tuned by training three independent Convolutional Neural Networks (CNNs). CNNs are chosen in preference to Recurrent Neural Networks (RNNs) in dealing with the potential multiple aspects discussed in each review~\cite{khan2018,Ren:2017:NND:3043977.3044107}. Then, to address the \textit{TransE} limitations using HIN, we model a reviewer as a separate node, rather than a connection between a product and a review. A graph inductive learning algorithm~\cite{Hamilton:2017:IRL:3294771.3294869} is used to fine-tune the pre-trained embeddings from the CNNs, which are concatenated by the respective Negative Ratio (NR) value (See Sec. \ref{sec:forward-impression}). Negative Ratio (NR), the proportion of a user's negative ratio is chosen, because it has been demonstrated as one of the most important user behavioral indicators in previous studies \cite{Shebuit2015,Shehnepoor2017}.
Other features used in \cite{wang2017handling,you2018attribute} such as Maximum Content Similarity (MCS) and Review Length (RL), are shown to be less significant in performance gain despite their higher computational cost \cite{Shehnepoor2017}. 
The benefit of the graph inductive learning is twofold: \textbf{first,} to facilitate the generation of embeddings for a new node, or a new (sub)graph in real-time;
and \textbf{second,} to refine the pre-trained embedding. 
DFraud$^3$ leverages component (review, user, and item) features, such as text features, metadata features, and also the graph structure (e.g., node degree), enabling the approach to learn an embedding function that generalizes the embedding features to unseen components. 
So every time a component is added, the inductive learning propagates the information to learn the component representation.
Finally, the representation is fed to a softmax layer for the final classification. 
Softmax is chosen over SVM, due to its ability to discriminate between samples with similar representation and different labels; a common case in fraud review detection with similar text for fraud and real reviews. 
In addition to the substantial performance gain (15\% for AUC) as compared to the state-of-the-art, our contributions of this work can be summarized as follows:

\begin{itemize}
\item 
We propose a novel three staged framework to address the cold-start problem using multi-component classification. This approach takes advantage of an HIN which considers item, review, and user as separate components. We employ SoWE to obtain a unique representation for each component, shown to be the first important performance contributor.
See Sec. \ref{subSubSec:performance}, \ref{sec:comp-anls}. 
\item 
For the first time, we propose a graph based inductive learning model for fraud detection that aggregates information of a node's neighborhood into a dense vector embedding, addressing the limitation of \textit{TransE} for multi-relation representation. 
Our extensive study demonstrates that graph based inductive learning is the second most important performance contributor, right after the CNN pre-trained component vectors. See Sec. \ref{sec:forward-impression}. 
\item
We investigate the camouflage problem as pointed out but not investigated in  \cite{Shebuit2015,Hooi:2017:GFD:3119906.3056563,Hooi:2016:FBG:2939672.2939747,Shehnepoor2017} when it occurs together with the cold-start problem. We devise a new approach to evaluate the performance of the system when facing the camouflage problem. Experimental results demonstrate that the DFraud$^3$ improves the detection of fraudsters who employ camouflage, with an increase in performance of 17\% as measured by AUC 
(see Sec. \ref{sec:Cam-anls}). 
\end{itemize}
The rest of the paper is structured as follows. In Section \ref{sec:rel-works}, we present the related work. In Section \ref{sec:method}, we introduce our methodology. In Section \ref{sec:results}, we show the experimental evaluation. We conclude the paper with an outlook to future work in Section \ref{sec:conclusion}.

\section{Related Works}
\label{sec:rel-works}
\subsection{The cold-start Problem}
Despite its significance, since the first work on Fraud Review Detection ~\cite{Jindal2008}, only a few studies investigated the cold-start problem. In particular, \cite{wang2017handling} employed three behavior features, namely,  Review Length (RL), Reviewer Deviation (RD), and Maximum Content Similarity (MCS) for fraud review detection. 

To mitigate the lack of information about a new user, i.e., the cold-start problem,  Wang \textit{et al.} ~\cite{wang2017handling} employed \textit{TransE}~\cite{Bordes:2013:TEM:2999792.2999923} to encode a graph structure between an item, a user and a review, where an item and a review are the head and the tail of a triple respectively, and the user who wrote the review for the item is considered as the relation. 
To learn vector representations of the three components, a training objective of \textit{TransE} is to minimize the distance between an $item$ vector after being translated by a $user$ vector in the embedding space and that of the $review$. An item's and a user's vector are randomly initialized from a random uniform distribution, while the embedding of a review is learned through a CNN, initialized using a pre-trained Word2Vec word embedding (CBOW)~\cite{Miklov:2013}. 
Results on the Yelp dataset show an accuracy of 65\%.

AEDA (Attribute Enhanced Domain Adaptive) is an attribute based framework proposed by \textit{You et al.} \cite{you2018attribute} to adapt the \textit{TransE} model from \cite{wang2017handling}. AEDA relies on the same concepts of that users are relations between items and reviews as in \cite{wang2017handling} 
to solve the cold-start problem. 
Three types of relationships are therefore defined, attribute-attribute, entity-attribute, and entity-entity between entities (review, item, and users). 
Different pairwise features (comparing two attributes of each entity) such as date difference (dateDif), rating difference (rateDiff) between two reviews are calculated for each entity as input for \textit{TransE}.
The results of the proposed framework shows a 75.4\% for accuracy on the Yelp restaurant dataset and 80.0\% on hotels, with an increase of 14\% as compared with \cite{wang2017handling}.

\subsection{Network-Based Fraud Detection}
As mentioned in Sec. \ref{sec:intro}, HIN as one of the network based models, has shown to be effective in network modeling \cite{Sun2012,Yu2012,Li2016}. There are also attempts on using network based approaches for fraud review detection, but they overlooked the cold-start.  

REV2~\cite{kumar2018rev2} formulates the fraudster detection as a bipartite network between users and products, and uses a Bayesian Inference Network (BIN) to iteratively learn the latent scoring about the fairness of reviews, quality of products, and reliability of reviewers. The performance is evaluated on 5 different datasets including Flipkart, Bitcoin OTC, Bitcoin Alpha, Epinions, and Amazon. It uses Laplacian smoothing to handle fraudster detection. 
Despite REV2 providing a theoretical guarantee for the performance with a 64.89\% for accuracy on fraudster classification, the approach does not perform well on the Yelp datasets. This is because in Yelp each user has only a single or a small number of reviews, resulting in a sparse network. 

Netspam~\cite{Shehnepoor2017} modeled fraud detection as a single component classification problem for fraud review detection. Features are extracted from text and metadata, and a \textit{metapath} is used to model the connection between every two reviews. Reviews are then labeled based on their similarity, through unsupervised and semi-supervised learning. Camouflage is discussed, and the impact of using the metapath is elaborated based on metapath weighting concept. However, no analytic explanation is provided to show how the framework works in face of camouflage.

SPeagle~\cite{Shebuit2015}, first extracts a vector of features from both text and metadata, then applies a function on the whole vector to calculate prior knowledge for fraudster group detection. For classification, Loopy Belief Propagation (LBP) is used. The results show significant performance on fraud detection on the Yelp dataset. Similar to Netspam, SPeagle also considers the possibility that a user might be a camouflaged fraudster. However, there is no discussion on how the framework performs in the face of camouflage.

\section{Proposed Method}
\label{sec:method}
\begin{figure}[ht]
  \includegraphics[width=\linewidth]{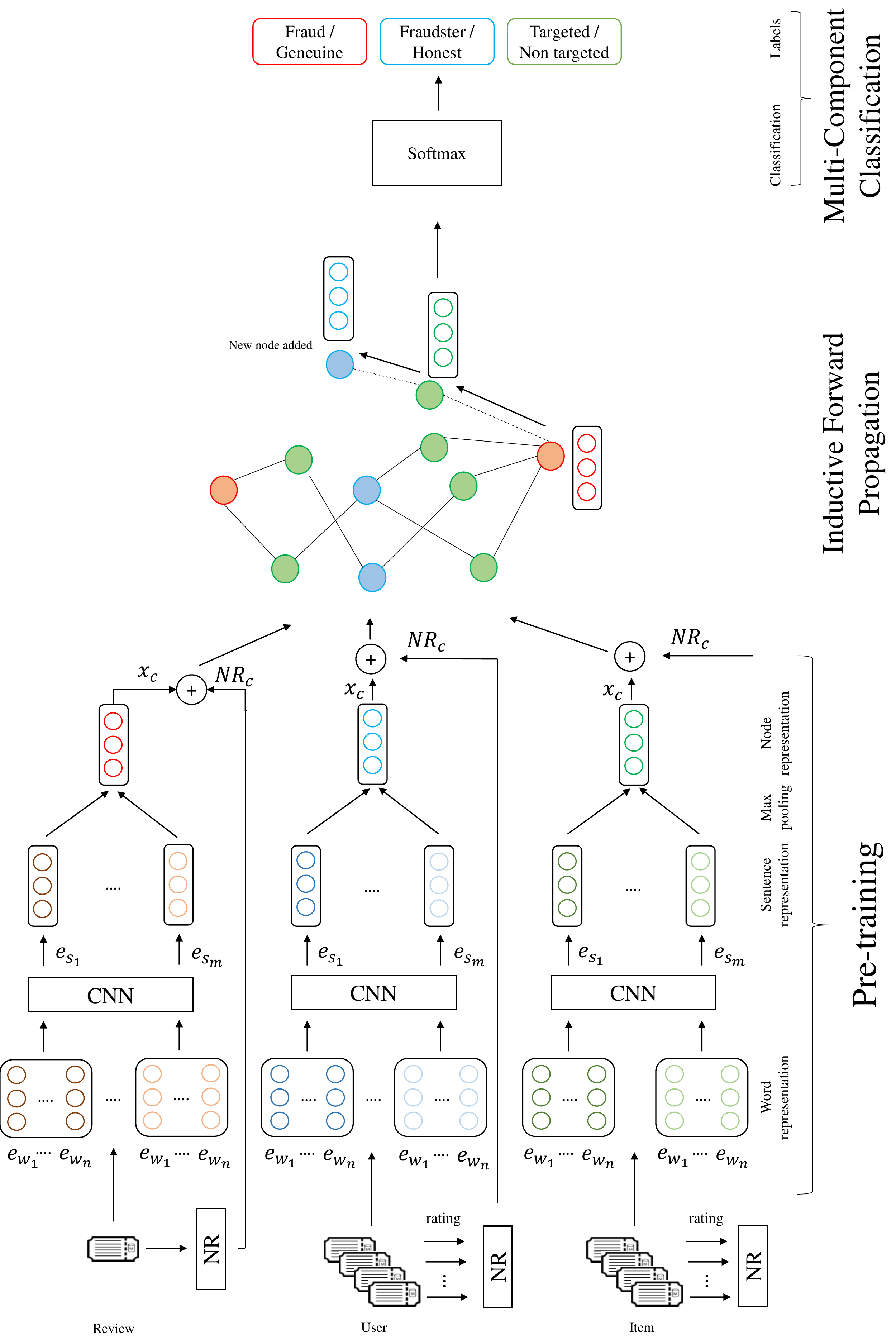}
  \caption{Framework of our proposed system.}
    \label{fig:framework}
\end{figure}

In this research, we propose to model a social review platform as a heterogeneous network, where each node is either a user, an item, or a review. The connections indicate a user has written a review for an item. Our proposed methodology follows three main steps as illustrated in Fig. \ref{fig:framework}. First, a vector representation for each component of the HIN including item, review, and user are obtained. For each item and user, reviews are aggregated and regarded as one document.The vector representation of each (aggregated) document is fine-tuned through a CNN. This text based representation obtained for components is then combined with the Negative Ratio (NR) as a behavioral feature. 
Next, This combination is then fed as an input to the inductive forward propagation of the HIN. 
Finally, a softmax layer is applied for a final multi-component classification. Fig. \ref{fig:framework} shows the overall framework of the DFraud$^3$. 

\subsubsection{Definition 1} 
(\textit{Multi Component Labeling}) Assume a graph $G = (U,I,R,E)$, where there are $N$ user nodes $U = \{u_1, ...,u_N\}$, $M$ item nodes $I = \{i_1, ..., i_M\}$, and $P$ review nodes $R = \{r_1, ..., r_P\}$ connected through edges $E$. The edge $E$ reflects two types of relations in the network; the edge between user and review $(u_n,r_p, type = ``write" )\in E$, and edge between review and item $(r_p,u_m, type = ``belong" )\in E$. The goal is to label each component in the graph. For each user, $L_U = \{fraudster, honest\}$, each item, $L_I = \{targeted, non-targeted\}$, and each review, $L_R = \{fraud, genuine\}$.

\begin{figure*}
  \centering
  \includegraphics[width=\linewidth]{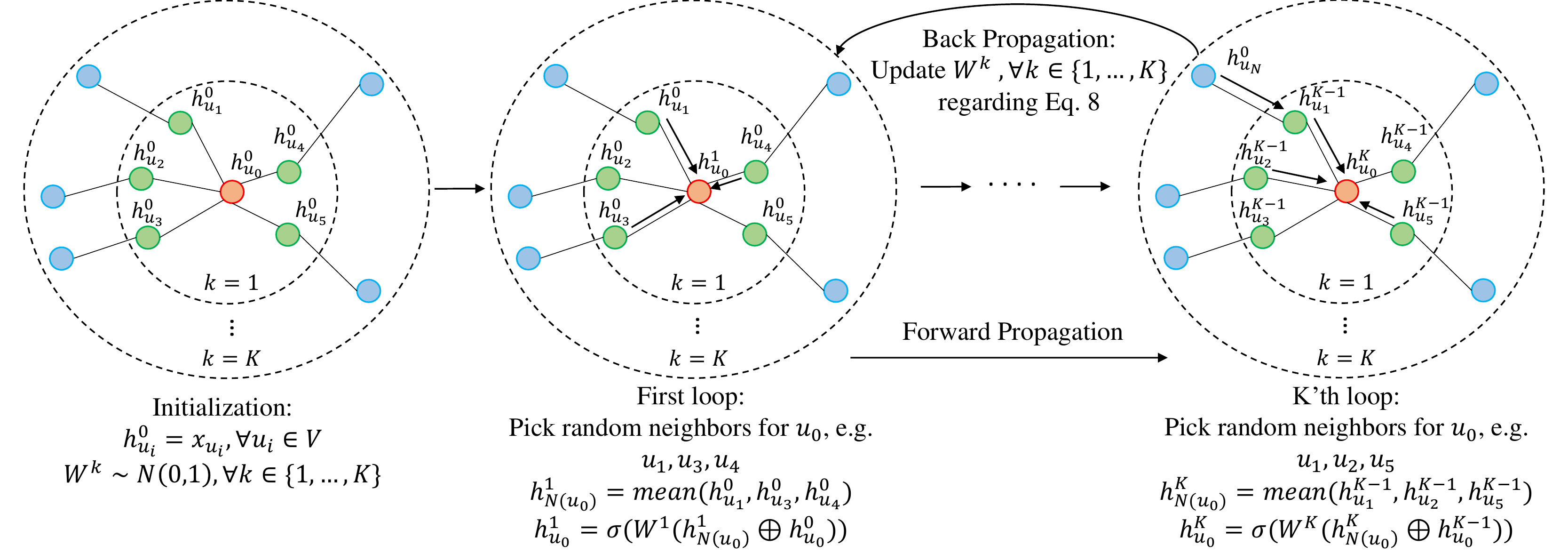}
  \caption{Toy example of our proposed graph learning (in this example $v = u_0$).}
    \label{fig:forward-prop}
\end{figure*}

\subsection{Pre-training}
\label{sec:Embedding-Extraction}
In the pre-training stage, we aim to learn an initial vector representation of each of the three components. 
Collective Intelligence (CI) \cite{Malone2010,malone-harnessing} states that the intelligence about a subject matter from a group, crowd and generally people about a subject, when considered together is a suitable representation of that subject matter. In the same token, we treat the aggregation of written reviews for a specific item as a suitable descriptor for that item. When consumers consider an item, they will look through all reviews and disregard who the reviewers are. So a natural representation for an item would be the collection of reviews. Similarly, the online footprint of a person can be an important behavior or character indicator for recruiting agencies. In other words, it makes sense to look through reviewers' comments collectively to gauge his/her reviews' behavior. Given its effectiveness and simplicity in a multitude of semantic-centered tasks \cite{Lyndon2015,6e8652a305e44e74ac5c829ea332e491}, the SoWE is adopted as an algorithm to obtain vector representations for each component. In other words, the SoWE of all reviews for an item is used as an initial vector representation for an item; the SoWE of all reviews written by a reviewer 
is used as the representation for the reviewer; and the SoWE of the tokens in a review as the initial vector for the review. These aggregated representations are much more meaningful covering global characteristics of each item and user, as compared with the random initialization, as which we will demonstrate in our results.
DFraud$^3$ includes three main sub-steps; word representation, sentence representation, and finally the node representation. 
\subsubsection{Word Representation}
\label{sec:word-represntation}
For a sentence containing $n$ words, we denote each word as $\{w_1, ...,w_n\}$, where the word $i$ embedding is represented as $e_{w_i}\in \mathcal{R}^D$, with $D$ as the word vector dimension. To obtain the representation, a look-up matrix, say $E$, is used, where $E\in \mathcal{R}^{D\times V}$, with $V$ the vocabulary size. Here, $E$ is initialized with a pre-trained word embedding~\cite{socher-etal-2013-recursive}.

\subsubsection{Sentence Representation}
\label{sec:sen-representation}
After pre-training the word embedding, a sentence model is trained using a shared CNN separately for each component (as shown in Fig.~\ref{fig:framework}). Inspired by \cite{Ren:2017:NND:3043977.3044107}, the CNN is trained in a supervised setting with the ground truth data as labels, to give a primary representation of the sentences. The convolutional layer in the CNN performs the role of a language model. 
The input for this layer is the concatenation of different words comprising the sentence, fed to a linear layer in a fixed-length window size equal to 3,  representing a trigram language model for the words. The concatenated word representations are denoted as $I_{3,i}\in \mathcal{R}^{D\times 3}$ where $D$ is the dimensionality of the word embeddings. The output of the linear layer is:
\begin{equation}
    \label{eq:LM-covolution}
    H_{i} = WI_{3,i} + b
\end{equation}
In Eq.~\ref{eq:LM-covolution}, $W\in \mathcal{R}^{L\times D\times D_3}$ ($L$ as output size of linear layer), and $b$ are shared parameters of the layer. 
Next, the output of the previous part is fed to an average pooling.
\begin{equation}
    \label{eq:ave-pooling}
    H = \frac{1}{n}\sum_{i = 1}^{n}H_i
\end{equation}
where $n$ is the number of words in a sentence.
Finally, a hyperbolic function $tanh$ is applied to incorporate non-linearity 
and obtain the final sentences' representation as follows:
\begin{equation}
    \label{eq:CNN-out}
    e_s = \tanh{H}
\end{equation}
where $e_s$ is the final embedding representation of the given sentence $s$ as input. The $e_s$ is then, fed to a softmax layer for classification. We use $e_s$ as the output. 

\subsubsection{Component Representation}
\label{sec:node-rep}
In this step, the input is the embedding for each sentence, required to be concatenated for each review, user, and item. 
\begin{equation}
    \label{eq:sentence-concat}
        e_c = e_{s_1} \oplus e_{s_2} \oplus ... \oplus e_{s_m}
\end{equation}
In Eq. \ref{eq:sentence-concat}, $e_c$ is the representation of component $c$, and $e_{s_i}\forall i\in m$ indicates the  representation of sentence $i$, and $m$ is the total number of sentences for $c$. A max pooling layer is applied to the input to obtain the representation for each unique component, as a node in the graph:
\begin{equation}
    \label{eq:max-pooling}
        x_c = \text{Max-pooling($e_c$)} 
\end{equation}
In Eq. \ref{eq:max-pooling}, $x_c$ is the final representation of component $c$. 
For an improved representation of each component, first the NR (Negative Ratio) for each user and item is calculated by following equation: 
\begin{equation}
    \label{eq:NR}
        NR_c = \frac{N(r = 1,2)}{N}
\end{equation}
In Eq. \ref{eq:NR}, $N(r)$ is the number of reviews with specific ratings ($r$) in range of 1-5 (5 is the highest), and $N$ is the total number of reviews for each component. 
To work out the NR for each type of components, we count the number of low ratings ($r = \{1,2\}$), divided by the number of total reviews for each component. 
Then NR is concatenated with the representation to obtain the final representation for each component in network:
\begin{equation}
    \label{eq:node-rep}
        x_v = x_c \oplus NR_c
\end{equation}
where $x_v$ in Eq. \ref{eq:node-rep} is a pre-trained feature representation for each component as node $v$ in the graph. 

Note that a new user is not introduced to the network unless he/she makes a review about an item. Once the review is written, the new user will be added to the network alongside the review, and the review is connected to an item, and this process continues. Items, regularly have connections in real-world datasets, making it easy to gather data from other reviews and users.

\subsection{Inductive Forward Propagation}
\label{sec:Embedding-generation}

\subsubsection{Objective Function}
\label{sec:obj-fun}
With pre-trained vectors as an input, for obtaining final graph based embeddings, an objective function is required to guarantee the satisfaction of two criteria: \textbf{(1)} neighbor nodes should have a similar representation, and \textbf{(2)} distant nodes should be apart in the embedding space. 
To satisfy these two criteria, we developed an unsupervised algorithm to learn the representations. Let $z_u,z_v$ be the final vector representation of vertex $u,v\in V$, respectively, where $v$ is in $u$'s neighbourhood, The objective function below employ Stochastic Gradient Descent (SGD) for training the weights:
\begin{equation}
    \label{eq:obj-fun-rep}
    J(z_u) = -\log(\sigma(z_u^Tz_v)) - Q\cdot E_{v_n\sim P_n(v)}\log(\sigma(-z_u^Tz_{v_n}))
\end{equation}
where $v$ is a node that connects with node $u$ in a specific neighborhood, with a predefined search depth of $K$, $\sigma$ is the sigmoid function. In addition, $P_n$ is a probability function for negative sampling, and $Q$ is the number of negative samples. The first term is to ensure that two similar nodes are close to each other in the embedding space. The second term ensures that negative samples, i.e., nodes that are not in the neighborhood of each other, should be distant from each other in the embedding space.
 
\subsubsection{Forward Propagation}
\label{sec:forward-prop}
We assume that the model is trained based on the objective function in Eq.~\ref{eq:obj-fun-rep} and with fixed hyper-parameters, namely $K, Q, H$, where $K$ is the specified maximum search depth, $Q$ is the number of negative samples, and $H$ is number of randomly selected neighbors. Intuitively, the reason for sampling is to reduce the computational complexity. The fixed size sampling is also to keep the computational cost for each batch fixed. Without using sampling, we will not be able to predict the memory used by each batch and the runtime of batch processing, which is $O(V)$ in the worst case, where $V$ is total number of nodes in the graph. On the other hand, per-batch space and time complexity would be fixed by the size of each batch. The testing process is thus: when a review from a new user is added to the system, $K$ aggregator functions (in this case the mean aggregator) are used to aggregate information from neighbors, with $K$ different weighting matrices known as $W^k,\forall k\in\{1,...,K\}$. The algorithm for the whole framework is described in Alg. \ref{alg:overall-alg}.

The key idea of Alg. \ref{alg:overall-alg} is that through each iteration of $k$, outer loop nodes' representations are combined with the neighbors' representations gradually.
As a result, in every iteration of $k$, the node's representation is combined with neighbors of one more depth, where $k$ represents the search depth. Note that $h_v^k$ denotes the node $v$ representation at depth $k$ and is initialized with the pre-trained features. 
In other words, for the first loop, $k = 0$, the representation $x_v$ is the pre-trained features from Sec. \ref{sec:Embedding-Extraction}, given as an input to the forward propagation system:

\begin{equation}
    \label{eq:h-init}
    h_v^0 = x_v
\end{equation}
Each iteration in the inner loop follows three main steps; \textbf{first,} the representations of a set of randomly selected neighbor nodes, $\{h_u^{k-1},\forall u\in N(v)\}$ are aggregated using  the ``mean" function, $h_{N(v)}^{k}$, which is a single vector indicating the aggregated values form the neighborhood. In every iteration, $h_{N(v)}^{k}$ is determined by the previous neighbor nodes' representations:
\begin{equation}
    \label{eq:neighbot-rep}
        h_{N(v)}^k = \frac{1}{M}\sum_{u}h_u^{k-1},\forall u\in N(v)\
\end{equation}
where $M$ is the number of randomly selected neighbor nodes of node $v$. 
Neighbor nodes are selected from a uniform distribution with probability less than 0.5.
\textbf{Next,} this vector is concatenated ($\oplus$) with the node's current representation, $h_v^{k-1}$ and then the resulting vector is fed to a fully connected layer with sigmoid ($\sigma$) as its activation function.
\begin{equation}
    \label{eq:k-rep}
        h_v^k = \sigma\left(W^k.(h_v^{k-1}\oplus h_{N(v)}^k)\right)
\end{equation}
In Eq. \ref{eq:k-rep}, $W^k$ is the weight matrix in the $k^{th}$ iteration. 
\textbf{Finally} the representation is normalized for each node $v$:
\begin{equation}
    \label{eq:norm-rep}
        h_v^k = h_v^k/||h_v^k||_2,\forall v\in V
\end{equation}
After $K$ steps, the generated representation $h_v^k$, is considered as the final representation of each node, $z_v$. These representations are then fed to a softmax layer for classifying each node. 
Note that the outputs of the softmax layer are the final classifications for each type of component, which makes the approach capable of multi-component classification.
The forward propagation and the training process through back propagation is depicted in Fig. \ref{fig:forward-prop}. 

\begin{algorithm}
 \caption{Proposed Algorithm}
\label{alg:overall-alg}
\textbf{Output:}\
Probability of new user to be fraudster (can be applied to review and item, as well)\;
\textbf{Input:}\
 Review $R(u,i)$ from new user $u$ on item $i$ with $S$ sentences and rating $r$\;

 \% Pre-training\;
 \For{$s \gets 1$ to $S$} 
    {
        $\{w_1, w_2, ..., w_n\} \leftarrow tokenize(s)$\;
        $\{e_{w_1}, e_{w_2}, ..., e_{w_n}\} \leftarrow CBOW(\{w_1, w_2, ..., w_n\})$\;
        $e_{s} \leftarrow CNN(\{e_{w_1}, e_{w_2}, ..., e_{w_n}\})$
    }
 \% New user representation\;
$r_c \leftarrow maxPooling(concat({e_{s},\forall s\in R(u,i)}))$\;
$x_v \leftarrow r_c \oplus NR_c$\;

 \% Forward-Propagation\;
 $h_{v}^{0} \leftarrow x_v,\forall v\in V$\;
  \%outerloop\;
  \For{$k \gets 1$ to $K$} 
    {\%innerloop\;        
    \For{$v\gets 1$ to $V$}
        {
            $h_{N(v)}^k \leftarrow mean(\{h_u^{k-1},\forall u\in N(v)\})$\;
            $h_v^k \leftarrow \sigma\left(W^k.(h_v^{k-1}\oplus h_{N(v)}^k)\right)$\;
        }
    $h_v^k \leftarrow h_v^k/||h_v^k||_2,\forall v\in V$\ \%normalization\;
}
 \% New user graph based representation\;
$z_v \leftarrow h_v^K,\forall v\in V$\;
\% Classification\;
$label(v) \leftarrow softmax(z_v)$

\end{algorithm}

\section{Results and Evaluation}
\label{sec:results}
\subsection{Datasets}
\label{sec:datasets}
To address the cold-start problem we require activity history provided through time stamps.
This will help us identify new users. 
Thus, for this research, we use time-stamped dataset\footnote{\url{http: //shebuti.com/collective-opinion-spam-detection/}} Yelp. Yelp is an online platform for people to share their experience of hotel and restaurant services in NewYork City (NYC). Other datasets such as TripAdvisor and Amazon lack either the ground-truth or timestamp. Hence they are not suitable in assessing the cold-start problem. Accordingly, similar to \cite{wang2017handling,you2018attribute} the state-of-the-art works on the cold-start, which we use as baselines for comparison, we conduct the experiments on the Yelp dataset. We prepared two subsets of data from the Yelp dataset to evaluate the performance of DFraud$^3$. The first one is Yelp-partial with randomly selected reviews from the whole dataset, The other is Yelp-whole which is the whole dataset containing all the reviews. Reviews in the datasets are labeled by the Yelp filtering system \cite{Shebuit2015}.
Table \ref{tab:datasets} summarizes the the two datasets.  

\begin{table} [ht]
\centering
  \caption{Basic statistics of the datasets.}
  \label{tab:datasets}
  \begin{tabular}{|c|ccc|}
    \hline
    Datasets & Reviews (fraud\%) & Users & Items \\ \hline
    Yelp-partial & 6,000 (17.34\%) & 4,046 & 46 \\
    Yelp-whole & 608,598 (13\%) & 260,277 & 5,044 \\ 
    \hline
    \end{tabular}
\end{table}

\subsection{Experimental Setup}
\label{sec:exp-set}
\subsubsection{System Setting}
\label{sec:exp-word-rep}
Recent years have seen more advanced word embedding techniques (ELMO \cite{peters-etal-2018-deep}, BERT \cite{devlin-etal-2019-bert}, XLNet \cite{DBLP:journals/corr/abs-1906-08237}) developed to refresh the new state of the art techniques on many natural language processing tasks. However, to provide the fair comparison with the two baseline systems (Wang \textit{et al.} \cite{wang2017handling} and You \textit{et al.} \cite{you2018attribute}, introduced in Sec. \ref{sec:rel-works}), we use the same embedding techniques i.e.  word embedding initialized using 100-dimension ($D$) Continuous Bag of Words (CBOW) \cite{journals/corr/abs-1301-3781} trained on Yelp dataset with a window size of 2. The vocabulary ($V$) size is $37,257$ for Yelp-partial, and $5,354,252$ for Yelp-whole. The learning rate is 0.1; batch size was set to 256, with 10,000 training epochs. For pre-training the CNNs, as mentioned in Sec. \ref{sec:sen-representation}, the filter size is 3, learning rate was set to 0.1, cross-entropy function is used as an objective function, with 30 training epochs. The initial values for $W$ and $b$ are set randomly from a uniform distribution. 
For training the graph-based representation, the minibatch number is 512, the learning rate was set to 0.01, the number of training iterations was set to 30 , with 3 as search depth ($K$). DFraud$^3$ is implemented in Python using Tensorflow 1.13. 
\subsubsection{Training and Test}
To determine the training and test set for evaluating DFraud$^3$ performance on the cold-start problem, reviews are split into two datasets with 80\% of the first reviews (based on timestamp) as the training set and remaining as the test set. The statistics of the sets are shown in Table \ref{tbl:train-test-stats}.
\begin{table} [ht]
    \centering
  \caption{Date range of training and test set for datasets.}
  \label{tbl:train-test-stats}
  \begin{tabular}{|c|c|cc|}
    \hline
    \multicolumn{2}{|c|}{Datasets} & Start date & End date \\ \hline 
    \multirow{2}{*}{Yelp-partial} & Train & 2005-05-10 & 2014-04-15 \\
    {} & Test & 2014-04-15 & 2015-01-10 \\ \hline
    \multirow{2}{*}{Yelp-whole} & Train & 2004-10-20 & 2014-05-20 \\
    {} & Test & 2014-05-20 & 2015-01-10 \\
    \hline
    \end{tabular}
\end{table}

\subsubsection{Labeling Procedure}
\label{sec:labeling}
Baseline systems \cite{you2018attribute,wang2017handling}, provide no information on how the labels from reviews are leveraged to become the ground truth for users  (i.e., fraudster or not). Similarly, the near ground-truth labeling procedure of the Yelp datasets contains labels only for reviews, not for users (fraudster or honest), nor items (targeted or non-targeted). To address this problem, Shebuit \textit{et al.} \cite{Shebuit2015} considered a user with at least one fraud review as a fraudster. This type of labeling can lead to inaccurate results, due to the near ground truth labeling procedure of reviews. We used a simple probability assignment for each user and item based on the fraud review they write and written for, respectively. A user $u(i)$ is a fraudster 
with the probability of $\frac{n_{fu}}{n_u}$
where $n_{fu}$
is the number of fraud reviews written by user $u$,
and $n_u$
is the total number of reviews by the same user $u$.
If the calculated probability is higher than 0.5 user $u$ 
is considered as a fraudster, 
otherwise $u$ 
is labeled as honest. 
Similarly, an item $i$ is targeted with a probability $\frac{n_{fi}}{n_i}$, where $n_{fi}$ and $n_i$ are the number of fraud reviews written for item $i$, and the number of total reviews written for item $i$ , respectively.

Since there is no ground truth on the camouflage problem, we devised a new approach to provide the labels. We used the camouflage definition (Sec. \ref{sec:intro}) to measure the effectiveness of DFraud$^3$ to uncover camouflaged users.
In other words, we looked for users with both fraud and genuine reviews in datasets and labeled them as suspicious of camouflage. For Yelp-partial, 137 users have multiple reviews with only 2 users suspicious of camouflage. While in Yelp-whole there are 90,179 users with multiple reviews and 2,121 users are suspicious of camouflage. Therefore, the approach was evaluated on the Yelp-whole for measuring the performance on the camouflage task, where 905 (out of 2,121) users are considered as camouflaged users for the test set (from original training and test set in Table \ref{tbl:train-test-stats}) and remaining (1215 users) are considered as training.


\subsection{Evaluation Metrics}
For evaluation, we rank the fraudster probability for each user. Users with higher values are more probable to be a fraudster. We used three standard metrics to describe the performance: Area Under Curve ($AUC$), Average Precision ($AP$), and F-measure.  
\subsubsection{Area Under Curve}
For $AUC$ \cite{Shebuit2015}, integration of the area under the plot of True Positive Ratio ($TPR$) on the $x$-axis and False Positive Ratio ($FPR$) on the $y$-axis is calculated. Consider $A$ as a list of sorted users in descending order according to their probability to be a fraudster. If we consider $n_j$ is the number of fraudster (honest) users sorted before the user in index $j$, then $TPR$ ($FPR$) for index $j$ is $\frac{n_j}{f}$, where $f$ is the total number of fraudster (honest) users. The $AUC$ is calculated as follows:
\begin{equation}
AUC = \sum_{i = 2}^{N}(FPR(i) - FPR(i-1))*(TPR(i))
\end{equation}
where $N$ is the total number of reviews.
\subsubsection{Average Precision}
For $AP$ \cite{Shehnepoor2017,Shebuit2015}, we need to have a list of sorted users based on their probability to be a fraudster. If $I$ is a list of sorted user indices based on their probability and $M$ is the total number of fraudster users, then $AP$ is formalized by:
\begin{equation}
    AP = \sum_{i=1}^{M}\frac{i}{I(i)}
\end{equation}

\subsubsection{F-measure}
\label{ref:F-measure}
Also known as $F1$ \cite{wang2017handling}, uses two main strategies for measuring performance, Micro and Macro. The former uses all correct estimations for different classes and then calculates the measure, regarding collected estimations, while the latter calculates the measure for each class separately, and then average the values. Obviously, with imbalanced data, using micro measure seems legit, while for balanced data macro measure can also be useful. $F1$ is calculated as follows:
\begin{equation}
    \label{eq:F1}
    F1 = \frac{2*precision*recall}{precision + recall}
\end{equation}

\subsection{Main Results}
\subsubsection{Ablative Study}
\label{subSubSec:performance}
To investigate the effectiveness of our graph based inductive learning, we used different combination of graph based approaches with different classifiers:

\textbf{Pre-trained + TransE + SVM (the effectiveness of pre-trained features):} This set is similar to study on \cite{wang2017handling} and \cite{you2018attribute}, and only differs in the pre-trained features (see Sec. \ref{sec:Embedding-Extraction}). Here, Word Embeddings (WE) and Negative Ratio (NR) are used as pre-trained features.

\textbf{Pre-trained + Inductive + SVM (the effectiveness of inductive learning):} To show the effectiveness of DFraud$^3$, the \textit{TransE} model is replaced by our proposed inductive learning approach in Sec. \ref{sec:forward-prop}.

\textbf{Pre-trained + Inductive + softmax (the effectiveness of softmax):} To observe the effectiveness of using the softmax as the classifier, the SVM is replaced with a softmax classifier. 

Table \ref{tab:comparison} shows DFraud$^3$ outperforms the two baseline systems on Yelp-partial. The results suggest that inductive learning yields better performance for all metrics. DFraud$^3$ performs better in terms of F1-Micro, while the results for F1-Macro is less encouraging, we attribute this to the unbalanced distribution of different classes over the partial dataset. The $AP$ and $AUC$ for DFraud$^3$ also demonstrate better results in comparison with baseline frameworks. Surprisingly, the performance of our approach is significantly improved using the softmax as the classifier for $AP$. One reason could be because the SVM works better for samples close to the margins. When the margin criteria is satisfied for the SVM, it will output the results. In better words, the SVM fails to model the samples with high feature similarity and different labels. As a result, the SVM works better for distant samples. The objective of the SVM is to maximize the margin which means for metrics like $AP$, the SVM works better since anomalies will not affect the classifier. On the contrary, the softmax objective is to produce a high probability for the correct class and small changes in samples can have a large effect on its performance. This can result in a noticeable difference in $AP$. 
\begin{table*}[h]
\centering
 \caption{Performance of DFraud$^3$ on cold-start user classification  in comparison with {\cite{wang2017handling}} and {\cite{you2018attribute}} on both datasets.}
  \label{tab:comparison}
  \begin{tabular}{|c|cccc|cccc|}
    \hline
    \multirow{2}{*}{Framework} & \multicolumn{4}{c|}{Yelp-partial} & \multicolumn{4}{c|}{Yelp-whole} \\ 
    {} & F1-Micro & F1-Macro & AP & AUC & F1-Micro & F1-Macro & AP & AUC\\ \hline
    Wang \textit{et al.} & 0.6031 & 0.5321 & 0.3232 & 0.5805 & 0.675 & 0.622 & 0.2691 & 0.5838 \\
    You \textit{et al.}& 0.6858 & 0.6021 & 0.3546 & 0.6353 & 0.7153 & 0.6431 & 0.3018 & 0.6283 \\
    pre-trained + TransE + SVM & 0.7035 & 0.6476 & 0.3696 & 0.6527 & 0.7602 & 0.6592 & 0.3348 & 0.6549 \\
    pre-trained + inductive + SVM & 0.7407& 0.6255 & \textbf{0.4432} & 0.6600 & 0.8372 & 0.5913 & \textbf{0.4372} & 0.6913 \\
    DFraud$^3$ & \textbf{0.7750} & \textbf{0.6666} & 0.3642 & \textbf{0.7009} & \textbf{0.8385} & \textbf{0.7087} & 0.3734 & \textbf{0.7331}\\ \hline
\end{tabular}
\end{table*}

By replacing the pre-trained features employed in \cite{you2018attribute} and \cite{wang2017handling} and keeping the \textit{TransE} model and SVM classifier, we observe that the performance is improved for all metrics. This indicates that our proposed pre-trained features are effective in capturing the feature of the three components of a review platform. Substituting the \textit{TransE} model with inductive learning results in a further improvement for all metrics, as compared with the two baseline systems, but a drop in F1-Macro as compared to \textit{TransE} model. The classifier adjustment to softmax brings improvement for most metrics, apart from $AP$. 

The results for the Yelp-whole dataset are displayed in Table \ref{tab:comparison}. Obviously, the performance improves using all the data from the training set.
In addition, the results on the Yelp-whole dataset show that DFraud$^3$ outperforms the two previous baseline systems for all four metrics. Similar to Yelp-partial, the performance is boosted for $AP$. In addition, except for F1-Macro inductive learning outperforms the \textit{TransE} model. One can explain the reason for the reduction of F1-Macro by justifying the data imbalance. As it is shown in Table \ref{sec:datasets}, 17\% of of reviews in Yelp-partial and  13\% of reviews in Yelp-whole are labeled as fraud reviews. This indicates a considerable imbalance between the number of fraud reviews and genuine ones.
 

\subsubsection{Multi-Component Classification Analysis}
\label{sec:comp-anls}
\begin{figure}
  \centering
  \includegraphics[width=\linewidth]{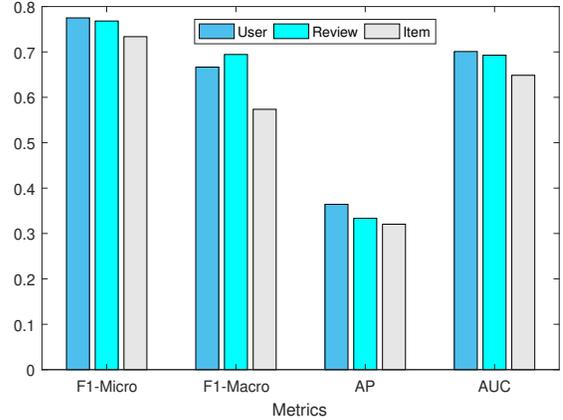}
  \caption{Comparison of DFraud$^3$ performance on different components for Yelp-partial (dark blue = user, light blue = review, gray = item).}
    \label{fig:partial-comp-anal}
\end{figure}

\begin{figure}
 \centering
 \includegraphics[width=\linewidth]{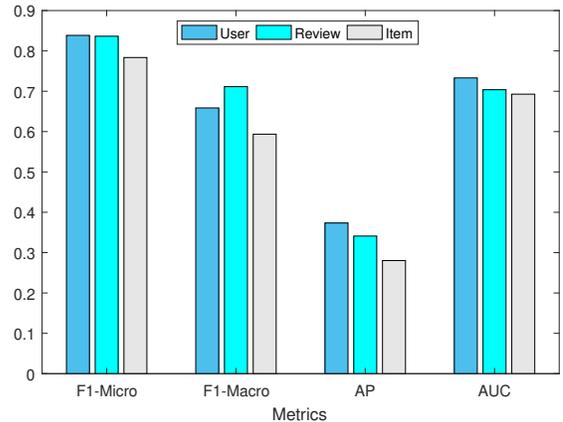}
 \caption{Comparison of DFraud$^3$ performance on different components for Yelp-whole (dark blue = user, light blue = review, gray = item).}
    \label{fig:whole-comp-anal}
\end{figure}

DFraud$^3$ performs classification on all of the three components. 
Fig. \ref{fig:partial-comp-anal}, \ref{fig:whole-comp-anal} depict the effectiveness of DFraud$^3$ for multi-component classification.
Results demonstrate that our system yields better performance on fraudster/honest user classification as compared with the classification of reviews and items. Considering the probability for each node as ground-truth (explained in Sec. \ref{sec:labeling}), instead of using binary labeling, it assists the model to detect the fraudsters with high performance in comparison with other types of components. In addition, the performance improves with more data as training data increases. 
Observation on datasets suggests that performance on three components reaches stability on Yelp-whole compared to Yelp-partial. This indicates that with complete data the performance is improved for users, items, and reviews.

\subsubsection{Impact of Inductive Learning}
\label{sec:forward-impression}
Another key difference between DFraud$^3$ and baseline systems is the use of the forward propagation after the pre-training step which outputs a refined primary representation of each component.
To observe the forward propagation's impact on the performance of the approach, we devised four different feature combinations: 

\textbf{Rand + Inductive:} A random feature representation is generated and then fed to the inductive learning for the final representation. The final representation is then fed to the softmax layer for final classification.

\textbf{WE (Word Embedding) + Inductive:} The pre-trained representation for this category is based on the word embeddings (WE) only excluding the NR. This representation is then fed to the inductive learning for final representation. The final labeling is based on the softmax classification. 

\textbf{WE + NR:} Inductive learning is withdrawn for this part and the pre-trained features are directly fed to softmax layer for final classification. 

\textbf{WE + NR + Inductive:} This represents the whole system.

\begin{figure}
  \centering
  \includegraphics[width=\linewidth]{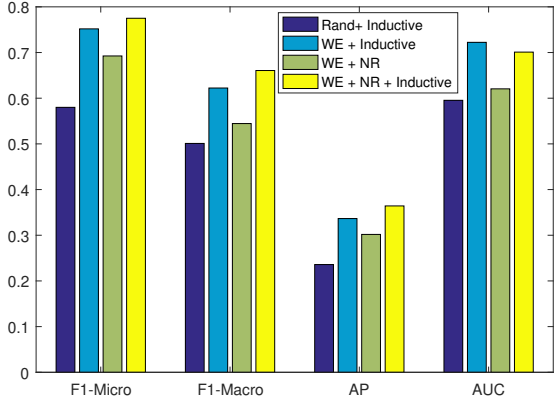}
  \caption{Impact of inductive learning on cold-start user classification on Yelp-partial 
  (purple = Rand + Inductive, blue = WE + Inductive, green = WE + NR, yellow = WE + NR + Inductive).}
    \label{fig:partial-feature-anal}
\end{figure}

\begin{figure}
 \centering
 \includegraphics[width=\linewidth]{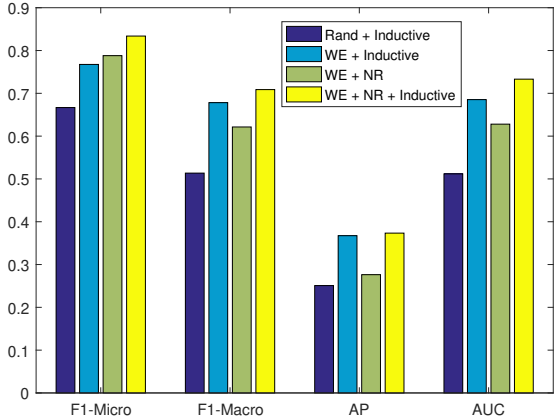}
 \caption{Impact of inductive learning on cold-start user classification on Yelp-whole (purple = Rand + Inductive, blue = WE + Inductive, green = WE + NR, yellow = WE + NR + Inductive).}
    \label{fig:whole-feature-anal}
\end{figure}

Fig. \ref{fig:partial-feature-anal}, \ref{fig:whole-feature-anal} represent the impact of inductive learning on both datasets regarding the mentioned metrics. There is a noticeable difference between the accuracy of the approach with and without the inductive learning (blue vs. green). Results demonstrate that the incremental aggregation of information from neighbors is effective at improving the system for addressing the cold-start problem (inductively based features outperform pre-trained features; yellow vs. blue). In addition, the performance of DFraud$^3$ on pre-trained features alone without graph embedding is already on par with baseline systems. In other words, even without applying inductive learning, DFraud$^3$ performs as well as previous studies. Furthermore, adding the NR feature improves the system performance which confirms previous works' findings \cite{Shebuit2015,Shehnepoor2017} regarding the importance of NR (yellow vs. blue). The only exception for the NR role in improving the performance is $AUC$ for the Yelp-partial dataset, which is due to a lack of information about all components in Yelp-partial. For F1-Micro, there is a small difference between inductive and NR (green vs. blue) which shows how much NR is effective in improving the performance of the approach. It is worth mentioning that with a bigger dataset the pre-trained features perform better than the small dataset (yellow vs. blue).


\subsubsection{Impact of N-1 Modelling}
\label{sec:N-1-handling}
\begin{figure}
 \centering
 \includegraphics[width=\linewidth]{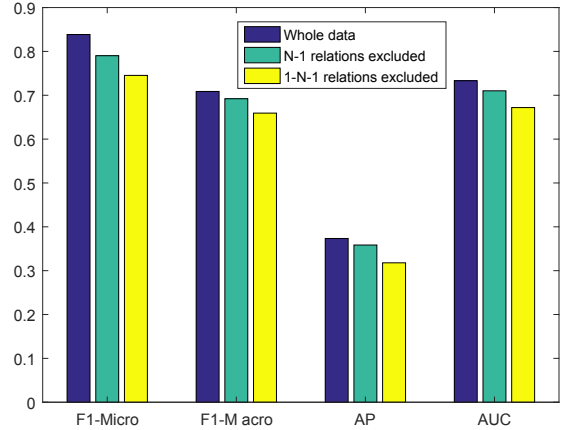}
 \caption{Impact of N-1, and 1-N-1 relations on cold-start user classification on Yelp-whole (purple = Whole data, green = N-1 relations excluded, yellow = 1-N-1 relations excluded).}
    \label{fig:N-1-anlysis-yelp-whole}
\end{figure}

As mentioned in Sec. \ref{sec:intro}, DFraud$^3$ handles the N-1 and 1-N-1 relations, which was the limitation of the \textit{TransE} model. To demonstrate that our performance gain is due to the better handling of N-1, and 1-N-1 relations, an experiment is conducted. In this experiment the N-1 relation, i.e., the same reviews written by the same users on different items; and 1-N-1, i.e. the same reviews written by different users, on the same item (\textit{TransE} ends up with the same representations for different users in this case) are 
removed. Fig. \ref{fig:N-1-anlysis-yelp-whole} represents the impact of N-1, and 1-N-1 relations removal on the performance.

As Fig. \ref{fig:N-1-anlysis-yelp-whole} shows, the removal of the same reviews by the same users on different items drops the performance for all measures. Intuitively, 
DFraud$^3$ makes use of user representation and its neighbors to calculate the final representation. More importantly, removing the relations with the same reviews on the same items with different users leads to a noticeable reduction performance as compared with the baseline systems. This, in turn, strengthened our claims that the performance gain of our system is due to its effectiveness in handling cold-start.
As a result of 1-N-1 removal, the system efficiency in handling cold-start is reduced, dramatically.

\subsubsection{Dealing with Camouflage}
\label{sec:Cam-anls}
As mentioned in Sec. \ref{sec:intro}, genuine reviews are not always written by honest people, and they can be written by fraudsters to hide their true identity. Previous approaches have not considered this problem since a fraudster can easily manipulate the traces by writing some honest reviews.  
We address this issue by using propagation over nodes. Then, the representation of each node is combined with its neighbors to regulate its importance in covering fraudsters. 

The performance on the two baseline systems is compared with DFraud$^3$. The results are presented in Table \ref{tab:camouflage-whole}.

\begin{table} [ht]
\centering
 \caption{Performance comparison on camouflaged users (1215 users as training set and 905 users as test set) in Yelp-whole 
 }
 \label{tab:camouflage-whole}
 \begin{tabular}{|c|cccc|}
    \hline
    Framework & F1-Micro & F1-Macro & AP & AUC \\ \hline
    Wang \textit{et al.}& 0.6212 & 0.6319 & 0.2801 & 0.5939 \\
    You \textit{et al.} & 0.6592 & 0.6602 & 0.2991 & 0.6082 \\
    DFraud$^3$ & \textbf{0.7846} & \textbf{0.6969} & \textbf{0.3940} & \textbf{0.7720}\\ \hline
\end{tabular}
\end{table}

\begin{figure}
 \centering
 \includegraphics[width=\linewidth]{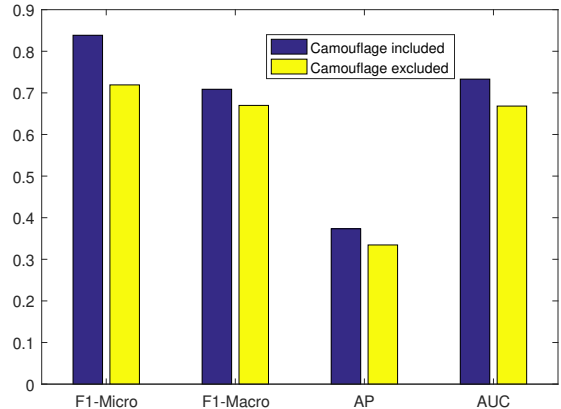}
 \caption{Impact of considering camouflaged users on DFraud$^3$ performance on Yelp-whole (purple = camouflaged users included, yellow = camouflaged users excluded).}
    \label{fig:camouflage-anlysis-yelp-whole}
\end{figure}

Table \ref{tab:camouflage-whole} presents the performance of the approach to camouflage detection against two baseline systems. We observe that our system outperforms the two baseline systems across all measures. Analysis suggests that using graph-based forward propagation helps the system to learn feature representations from neighbor nodes, which helps uncover the true intentions of users for writing contradicting reviews in terms of authenticity.
Similar to Sec. \ref{sec:N-1-handling}, we conducted an experiment to demonstrate that the gain in the performance is also due to the better handling the camouflage users. Fig. \ref{fig:camouflage-anlysis-yelp-whole} shows the performance of DFraud$^3$ for two cases: when camouflaged users are included, and when they are excluded from the dataset. As we can see, the performance drops after excluding the camouflaged users. Analytically, camouflaged users first write genuine reviews to hide their true intentions, in the worst-case scenario. This means that the fraud detection system requires information from both the neighbors and the node itself. Previous approaches employed information only from one-hop neighbors, which is not helpful in cases of camouflage. Also, they missed the opportunity of using the initial information for each user, which can be used to initialize the pre-knowledge of each node. To address the first problem, graph-based inductive learning facilitates information propagation for more than one hop and it helps to gather information from distant nodes rather than just the neighbor nodes. The second limitation is addressed using the pre-training step (Sec. \ref{sec:Embedding-Extraction}). This leads to the effective detection of camouflaged users.    

\section{Conclusion}
\label{sec:conclusion}
Cold-start is a challenging issue that hinders the effective detection of fraudsters in social review platforms. In this research, we devised a system that takes advantage of the textual and rating data (abundant surface data) 
and aggregates them through a CNN as initially learned features for a vector representation of each component of a social review platform. The initial vector representation is then refined through a graph inductive learning algorithm we proposed to capture the interplay between a user, an item and a review, using multi-component classification 
(reviews to fraud, genuine; user to fraudster, honest; and items to targeted, non-targeted) 
as the downstream task. 
Two sets of comprehensive ablative studies have been carried out that demonstrate the effectiveness of our approach to learning the representation of each component. 
Notably, there is significant performance gain achieved by WE + NR and performing inductive learning on the Yelp dataset from two domains; restaurants and hotels. Defining a new relationship between components, from a different view can be seen as future work. 
One way is to consider each link's importance regarding metapath weight \cite{Shehnepoor2017} to calculate contributions of the influence of each link in the final classification. This also can be applied to contents from other media such as twitter to assist spam detection \cite{Fazil2018,Yang2013,Ruan2016}. 
\bibliographystyle{IEEEtran}
\bibliography{References}

\end{document}